\documentclass[journal]{IEEEtran}
\usepackage{amsmath,amsfonts}
\usepackage{algorithmic}
\usepackage{algorithm}
\usepackage{array}
\usepackage[caption=false,font=footnotesize,labelfont=rm,textfont=rm]{subfig}
\usepackage{textcomp}
\usepackage{stfloats}
\usepackage{url}
\usepackage{verbatim}
\usepackage{graphicx}
\usepackage{cite}

\usepackage{multirow}
\usepackage{xcolor,color,framed}

\begin{document}

%\title{Large Language Model-enhanced Reinforcement Learning for Low-Altitude Economy Networking: A Tutorial and Case Study}

\title{Large Language Model-enhanced Reinforcement Learning for Low-Altitude Economy Networking}

\author{Lingyi Cai, Ruichen Zhang, Changyuan Zhao, Yu Zhang, Jiawen Kang,~\IEEEmembership{Senior Member,~IEEE},\\ Dusit Niyato,~\IEEEmembership{Fellow,~IEEE}, Tao Jiang,~\IEEEmembership{Fellow,~IEEE}, and Xuemin Shen,~\IEEEmembership{Fellow,~IEEE}

        % <-this % stops a space
\thanks{Lingyi Cai is with the Research Center of 6G Mobile Communications, School of Cyber Science and Engineering, Huazhong University of Science and Technology, Wuhan, 430074, China, and also with the College of Computing and Data Science, Nanyang Technological University, Singapore (e-mail: lingyicai@hust.edu.cn).}
\thanks{Ruichen Zhang, Changyuan Zhao, and Dusit Niyato are with the College of Computing and Data Science, Nanyang Technological University, Singapore (e-mails: ruichen.zhang@ntu.edu.sg; zhao0441@e.ntu.edu.sg; dniyato@ntu.edu.sg).}
\thanks{Yu Zhang and Tao Jiang are with the Research Center of 6G Mobile Communications, School of Cyber Science and Engineering, Huazhong University of Science and Technology, Wuhan, 430074, China (e-mail: yuzhang123@hust.edu.cn; tao.jiang@ieee.org).}
\thanks{Jiawen Kang is with the School of Automation, Guangdong University of Technology, Guangzhou 510006, China (e-mail: kavinkang@gdut.edu.cn).}
\thanks{Xuemin Shen is with the Department of Electrical and Computer Engineering, University of Waterloo, Waterloo, ON N2L 3G1, Canada (e-mail: sshen@uwaterloo.ca).}

}

%\markboth{Journal of \LaTeX\ Class Files,~Vol.~14, No.~8, August~2021}


\maketitle

\begin{abstract}
Low-Altitude Economic Networking (LAENet) aims to support diverse flying applications below 1,000 meters by deploying various aerial vehicles for flexible and cost-effective aerial networking. However, complex decision-making, resource constraints, and environmental uncertainty pose significant challenges to the development of the LAENet. Reinforcement learning (RL) offers a potential solution in response to these challenges but has limitations in generalization, reward design, and model stability. The emergence of large language models (LLMs) offers new opportunities for RL to mitigate these limitations. In this paper, we first present a tutorial about integrating LLMs into RL by using the capacities of generation, contextual understanding, and structured reasoning of LLMs. We then propose an LLM-enhanced RL framework for the LAENet in terms of serving the LLM as information processor, reward designer, decision-maker, and generator. Moreover, we conduct a case study by using LLMs to design a reward function to improve the learning performance of RL in the LAENet. Finally, we provide a conclusion and discuss future work.
\end{abstract}

\begin{IEEEkeywords}
Low-altitude economy networking, reinforcement learning, large language model, decision-making, reward design.
\end{IEEEkeywords}

\section{Introduction}

% 1.25页

%Introduce the concept of LAENet. Explain the need for intelligent orchestration and optimization algorithms in LAENet to adapt to dynamic environments and balance multiple objectives such as energy efficiency, communication coverage, and QoS.

%The concept of RL is introduced and its ability to solve the needs in LAENet is shown. However, RL has some limitations in LAENet.

%Introduce the concept of LLM and explain that LLM has the potential to solve the issues in RL.

%In this article, we propose ...

%10833672

Low-altitude economic networking (LAENet) refers to the integration of communication and network infrastructure designed to support the deployment of manned and unmanned aerial vehicles (UAVs) in the airspace below 1,000 meters \cite{10759668}. The primary goal is to generate commercial and societal value through diverse aerial operations. Specifically, the LAENet is distinguished by its high mobility, adaptive deployment, and cost-effectiveness \cite{10972017}. For example, these aerial platforms can serve as mobile and cost-effective communication nodes capable of acting as aerial base stations, communication relays, and edge computing devices \cite{10972017}. Thus, the LAENet can support diverse applications such as intelligent transportation, disaster response, and ubiquitous telecommunications. However, these advantages are accompanied by several significant challenges: real-time decision-making for UAV operation and network coordination; environmental uncertainties with unpredictable channel conditions and user mobility; and resource-constrained heterogeneity such as limited energy and communication capacity \cite{10759668}.

To address these challenges, reinforcement learning (RL) emerges as a promising solution for the LAENet \cite{9750860}. Specifically, RL enables autonomous and adaptive control, allowing aerial vehicles to make time-sensitive decisions without reliance on predefined models. By continuously observing and interacting with dynamic environments, RL can facilitate robust decision-making under uncertainty. Additionally, RL can learn optimized policies to manage resources under constraints of energy, bandwidth, and computational capacity. However, classical RL still has limitations in addressing the challenges faced by the LAENet \cite{10766898}. Classical RL often struggles with generalization, as models trained for specific tasks lack the capacity to adapt to new and dynamic scenarios. Importantly, the manual design of reward functions in classical RL can lead to suboptimal policies or unintended behaviors if not properly formulated. Furthermore, the RL methods are prone to performance degradation due to error accumulation in the decision-making process.

Recently, with the rise of large language models (LLMs) technology, enhancing classical RL by leveraging the strengths of LLMs as an integrated approach to address key challenges in the LAENet has emerged as a promising research direction. LLMs trained on massive and diverse datasets exhibit a strong capacity for generation, contextual understanding, and structured reasoning \cite{10766898,10679152}. For example, LLMs can perform cross-domain knowledge transfer to provide relatively accurate responses when facing varying inputs and cross-scenario tasks. Through their chain-of-thought reasoning and understanding, LLMs can construct effective signals for decision-making by capturing nuanced trade-offs and important objectives in the context of tasks \cite{NEURIPS2022_9d560961}. These properties make LLMs suitable for integration into RL that requires flexibility, adaptability, and policy learning under limited prior knowledge and an uncertain environment.

In this paper, we first provide a tutorial about utilizing LLMs to integrate into RL by reviewing existing related works, which shows the benefits of LLMs in improving the performance of classical RL. Then, we propose an LLM-enhanced RL framework in the context of the LAENet, where the LLM is used to interpret complex multimodal inputs, shape adaptive reward functions, to guide or generate action sequences, and simulate future state transitions in the learning process. In addition, we implement a case study on utilizing the LLM to shape the reward function of RL to optimize the energy consumption in the LAENet. Finally, we conclude this paper and explore the future research directions on LLM-enhanced RL for the LAENet. The main contributions of this paper are summarized as follows:

\begin{enumerate}
  \item We comprehensively analyze how LLMs can address the limitations of classical RL and synthesize existing research to guide future integration efforts, which provides a useful foundation for leveraging LLMs to enhance the capabilities of RL.

  \item We propose a novel LLM-enhanced RL framework for the LAENet, where the LLM functions as an information processor, reward designer, decision-maker, and simulator, effectively unifying the complementary strengths of LLMs and RL.

  \item We provide a case study to demonstrate the benefit of using LLMs to design reward functions for RL agents in the LAENet. Compared to manually crafted rewards, the LLM-designed reward leads to more efficient learning and improved performance.

\end{enumerate}

%\section{Background of RL and LLM}
\section{Overview of enhancing RL with LLM}

In this section, we comprehensively review the background knowledge of RL and LLM. Then, we highlight the potential of LLMs for enhancing RL.

%\cite{sutton1998reinforcement}

%\cite{kaelbling1996reinforcement,watkins1992q,mnih2013playing,mnih2015human,schulman2017proximal,haarnoja2018soft}

%\cite{vaswani2017attention,devlin2019bert,radford2018improving,radford2019language,brown2020language,touvron2023llama}

\subsection{Background of RL}

%\subsubsection{Classical RL} 
%RL is a foundational ML paradigm driven by a trial-and-error mechanism, where the agent observes the current state of the environment to select the action, and receives a reward signal that reflects the quality of the action \cite{10766898}. The agent’s objective is to maximize the cumulative long-term reward, often referred to as the return. A classical formulation of RL problems is the Markov Decision Process (MDP), typically described by the tuple $(S,A,P,R,\gamma )$, where $S$ and $A$ represent the sets of states and actions, $P$ defines the transition probability between states, $R$ specifies the reward structure, and $\gamma  \in [0,1]$ is the discount factor balancing immediate and future rewards \cite{10283826}. The agent iteratively updates its policy to improve its expected return, eventually converging toward an optimal or near-optimal solution. In this context, RL has the ability to enable aerial vehicles to learn optimal strategies through interaction with dynamic environments for real-time decision-making and autonomous coordination in the LAENet.

RL is a foundational ML paradigm driven by a trial-and-error mechanism, where the agent observes the current state of the environment to select the action, and receives a reward signal that reflects the quality of the action \cite{10766898}. The agent’s objective is to maximize the cumulative long-term reward, often referred to as the return. The expected return is improved by iteratively updating the agent's policy to eventually converge to an optimal or near-optimal solution \cite{kwon2023reward}. In this context, RL agents have the ability to learn optimal strategies through interaction with dynamic environments for real-time decision-making and autonomous coordination.

%In this context, RL has the ability to enable aerial vehicles to learn optimal strategies through interaction with dynamic environments for real-time decision-making and autonomous coordination in the LAENet \cite{10283826}.

%Thus, RL enables aerial vehicles in LAENet, either centrally or autonomously controlled, to learn optimal strategies through interaction with dynamic environments.

%\subsubsection{Limitations of Classical RL in LAENet} Explain the limitations of classical RL in LAENet, which can be enhanced using LLM.

\begin{figure*}[t]
\centering
\includegraphics[width=1\linewidth]{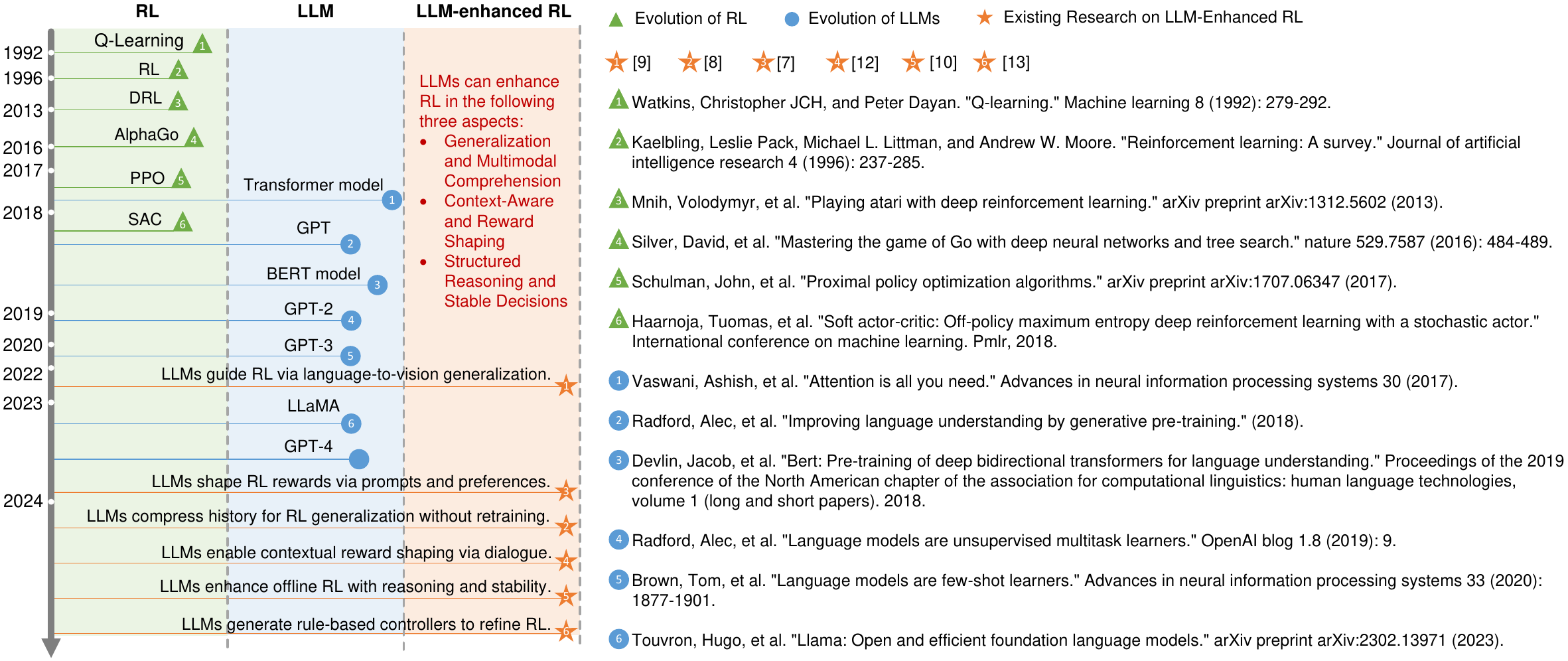}
\caption{An illustration of RL, LLM, and applications of LLM-enhanced RL. The number of peer-reviewed publications regarding RL, LLM, and LLM-enhanced RL per year is shown on the left-hand side (the publication data was collected from IEEE Xplore in April 2025).}
\label{paperlist-v2LLMRL}
\end{figure*}

RL has evolved from early model-free methods (such as Q-learning) to advanced DRL techniques capable of handling high-dimensional inputs. Key milestones include the introduction of Deep Q-Networks (DQN) and subsequent algorithms such as PPO and SAC for applying RL in complex environments, as shown in Fig. \ref{paperlist-v2LLMRL}. However, several challenges still limit the performance of RL.

%\subsubsection{Generalization and Multimodal Understanding} The LAENet operates in dynamic, heterogeneous, and uncertain environments, involving changing channel conditions, diverse signal metrics, and various spatial information. In such complex settings, classical RL trained in specific environments may be difficult to generalize to new deployment scenarios due to lacking the ability to process multimodal data.

\subsubsection{Generalization and Multimodal Understanding} In the dynamic and uncertain environments, classical RL trained in specific environments may be difficult to generalize to new and complex scenarios due to lacking the ability to process multimodal data (e.g., visual and language data for RL agents in robotics applications).

%The LAENet operates in dynamic and uncertain environments, involving changing channel conditions and various spatial information. In such complex settings, classical RL trained in specific environments may be difficult to generalize to new scenarios due to lacking the ability to process multimodal data.

\subsubsection{Reward Function Design and Feedback} It is challenging to define reward functions that trade off among multiple objectives in the RL. Inappropriately designed rewards can lead to suboptimal policy learning and unintended behaviors.
%It is challenging to define reward functions that trade off among multiple objectives in the LAENet, such as energy efficiency, latency minimization, and QoS. Inappropriately designed rewards can lead to suboptimal policy learning and unintended behaviors.

\subsubsection{Model Instability and Lack of Interpretability} Model-based RL suffers from error accumulation of models in dynamic environments. Moreover, the decision-making process of classical RL lacks interpretability and may be unsuitable for safety-critical scenarios.

%Model-based RL faces the challenge of errors in the model’s predictions accumulating due to variable dynamics of the LAENet. In addition, the decision-making process of classical RL is usually inexplicable, which may not meet the requirements of safety-critical applications in the LAENet.

%Model-based Planning Instability and Lack of Interpretability: Compounding model errors and lack of transparent decision logic limit the robustness and trustworthiness of RL agents.

%Sample Inefficiency and Multi-task Interference: RL demands large amounts of data and suffers from task interference in multi-task settings, reducing learning efficiency and transferability.

%\subsubsection{Sample Inefficiency and Multi-task Interference} classical RL requires extensive interaction with the environment and collecting sufficient training data to iterate optimal policies. Additionally, it can be affected by interference in managing multiple tasks of the LAENet.

LLMs are advanced deep learning models trained on extensive datasets at the terabyte-level and typically characterized by billions of parameters \cite{10766898,10679152}. As shown in Fig. \ref{paperlist-v2LLMRL}, the development of LLMs has been driven by the Transformer architecture that laid the foundation for bidirectional masked language modeling (e.g., BERT) to autoregressive generative pretraining (e.g., GPT series). With continued scaling of model size and data, the release of open-source foundation models such as LLaMA aims to reduce the number of parameters while maintaining model performance. Such massive scale of data and model complexity enables LLMs to achieve remarkable capabilities in language generation, knowledge representation, and logical reasoning. In this case, some favorable properties of LLMs have the potential to enhance classical RL to overcome the above limitations, as shown in Fig. \ref{paperlist-v2LLMRL}.

\subsection{Background of LLM and Its Potential to Enhance RL} %Provides background of LLMs that can be used to enhance RL, including definition of LLM, properties of LLM and its ability to enhance RL

%\subsubsection{Generalization and Multimodal Comprehension} {\color{blue}LLMs are pretrained on broad datasets and can process diverse data modalities, such as textual commands, spatial layouts, and raw visual inputs. These complex inputs may be difficult for classical RL agents to interpret and generalize due to reliance on domain-specific encoders.} To overcome these limitations, the authors in \cite{pmlr-v162-paischer22a} leveraged frozen LLMs as high-level planners to bridge abstract natural language instructions with motion control in robotic tasks, thereby enable generalization of RL agent using only raw visual observations without task-specific retraining. The experimental results\footnote{Source code available on: https://github.com/ml-jku/helm.\label{s1}} show that this proposed scheme outperformed Long Short-Term Memory (LSTM)-based baselines by 18.5\% in sample efficiency. Similarly, the study in \cite{dalal2024planseqlearn} utilized a frozen pre-trained language model to compress past observations into semantic representations, enabling RL agents to learn from historical context for generalization without retraining. Results\footnote{Source code available on: https://github.com/mihdalal/planseqlearn.\label{s2}} demonstrate that generalization is achieved across more than twenty-five tasks and outperforms the hierarchical RL baselines by 85\%.

\subsubsection{Generalization and Multimodal Comprehension} LLMs are pretrained on broad datasets and can process diverse data modalities, such as textual commands, spatial layouts, and raw visual inputs. These complex inputs may be difficult for classical RL agents to interpret and generalize due to reliance on domain-specific encoders. To overcome these limitations, the authors in \cite{pmlr-v162-paischer22a} leveraged frozen LLMs as high-level planners to bridge abstract natural language instructions with motion control in robotic tasks, thereby enable generalization of RL agent using only raw visual observations without task-specific retraining. The sample efficiency of the proposed scheme outperformed Long Short-Term Memory (LSTM)-based baselines by 18.5\%. Similarly, the study in \cite{dalal2024planseqlearn} utilized a frozen pre-trained language model to compress past observations into semantic representations, enabling agents to learn from historical context for generalization without retraining and to outperform the hierarchical RL baselines by 85\%.

\begin{table*}[htbp]
\centering
\caption{Summary of recent Projects on LLM-enhanced RL}
\renewcommand{\arraystretch}{1.3}
\begin{tabular}{|c|m{2.3cm}|m{1.8cm}|m{1.5cm}|m{3.2cm}|m{3.1cm}|c|}
\hline
\textbf{Related  Ref.} & \centering \textbf{GitHub Link} & \centering \textbf{Tasks Supported} & \centering \textbf{LLM type} & \centering \textbf{Training Data} & \centering \textbf{Performance Metrics} & \textbf{Last Update} \\
\hline
{\cite{pmlr-v162-paischer22a}} & \url{https://github.com/ml-jku/helm} & \centering Partially Observable RL & \centering Transformer-XL & \centering RandomMaze, Minigrid, Procgen & \centering Sample Efficiency, IQM & Dec. 2024 \\
\hline
{\cite{dalal2024planseqlearn}} & \url{https://github.com/mihdalal/planseqlearn} & \centering Long-horizon robot control & \centering GPT-4 & \centering Meta-World, ObstructedSuite, Kitchen, Robosuite & \centering Sample Efficiency, Task Success Rate & Aug. 2024 \\
\hline
{\cite{kwon2023reward}} & \url{https://github.com/minaek/reward_design_with_llms} & \centering Reward design in RL & \centering GPT-3 & \centering Ultimatum Game, Matrix Games, DEALORNODEAL & \centering Labeling Accuracy, User Alignment Score & May. 2023 \\
\hline
{\cite{shi2024unleashing}} & \url{https://github.com/srzer/LaMo-2023} & \centering Offline RL & \centering GPT-2 & \centering D4RL, d4rl-atari & \centering Sample Efficiency, Sparse Reward Performance & Jun. 2024 \\
\hline
{\cite{NEURIPS2023_1b44b878}} & \url{https://github.com/noahshinn/reflexion} & \centering Reasoning and Programming & \centering GPT, starchat-beta & \centering ALFWorld, HotPotQA, HumanEval, MBPP & \centering Pass@1 Accuracy, Exact Match, Hallucination Rate & Jan. 2025 \\
\hline
\end{tabular}
\end{table*}

\subsubsection{Context-Aware and Reward Shaping} LLMs can shape reward functions to balance multiple objectives or constraints for RL agents by using domain knowledge and context comprehension, which address the limitation of RL's static or unsuitable reward function that requires extensive domain expertise. For example, the work in \cite{kwon2023reward} shaped RL agent's reward model by leveraging LLMs to prompt examples and preference descriptions to guide agent behavior. The LLM-based rewards achieved 91\% accuracy in the Ultimatum Game versus 67\% for conventional reward engineering approaches. In a more complex application of active distribution networks, the study in \cite{yang2024rl2} utilized LLMs to enable context-aware reward shaping for RL agents by leveraging domain knowledge and iterative refinement through multi-round dialogues. As a result, the LLM-based reward-shaping approach reduced performance variance by 89.4\% over conventional fixed-reward methods.%, highlighting LLM's ability to adjust reward signals based on changing environments.  

%LLMs can shape reward functions to balance multiple objectives in the LAENet by using domain knowledge and context comprehension, which reduces the risk of suboptimal policies due to unsuitable rewards.

% LLMs may achieve more transparency in decision-making by supporting step-by-step and structured reasoning through prompting strategies such as Chain of Thought. Meanwhile, the decision-making in the LAENet can be more stable due to the capacity of LLMs on transferring knowledge across domains.

\subsubsection{Structured Reasoning and Stable Decisions} LLMs may achieve more transparency in decision-making for RL agents by supporting step-by-step and structured reasoning through prompting strategies such as Chain of Thought. Meanwhile, the decision-making in the RL can be more stable due to the capacity of LLMs in transferring knowledge across domains. For instance, the authors in \cite{shi2024unleashing} enhanced decision stability and structured reasoning in offline RL by leveraging pre-trained language models' sequential knowledge and linguistic representations. Compared to the value-based offline RL algorithm, this scheme reduces performance variance by 40\% even if the data is 1\% of the whole dataset. The work in \cite{10529514} utilized LLMs to generate structured and rule-based controllers through step-by-step prompting for robotic manipulation, where the RL benefits from the integration with these controllers to stabilize and refine policy learning. The proposed scheme maintains error rates below 0.12\% in dynamic manipulation tasks compared to 4.7\% of the TD3 algorithm baseline.

\iffalse
{\bf Lesson Learned}: Our review shows that the core RL mechanisms can be enhanced through integration with LLMs, as shown in Fig. \ref{paperlist-v2LLMRL}, which can extract abstract representations from multimodal inputs, guide reward design through contextual reasoning, and support structured policy generation via step-by-step prompting. These integrations enable RL agents to perform effectively across diverse tasks, sparse-reward settings, and dynamic environments without task-specific retraining. The potential of enhancing RL with advantages of LLMs can be further manifested in various applications.
\fi
%, especially in multi-agent systems such as the LAENet.

%\subsubsection{Sample Generation and Multi-task Management} LLMs can leverage their prior knowledge to generate data via simulation or in-context learning, thus reducing the complex interactions with the environments. Moreover, transferring broad understanding to various tasks helps manage multiple tasks simultaneously in the LAENet.

\begin{figure*}[t]
\centering
\includegraphics[width=1\linewidth]{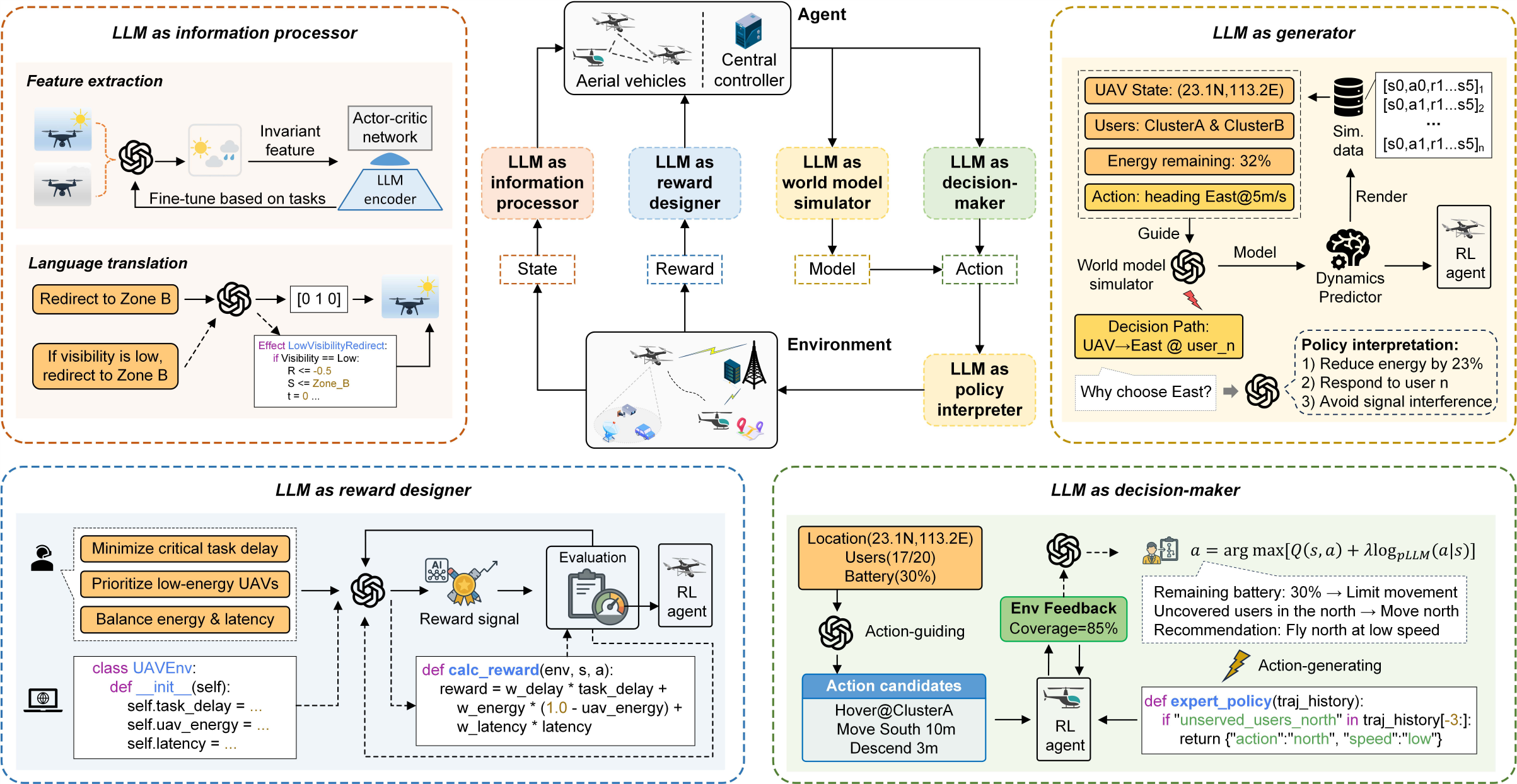}
\caption{An overview of the LLM’s multiple roles in reinforcement learning, including information processor, reward designer, decision-maker, and generator, highlighting its central role in bridging language input and decision-making processes within the LAENet framework.}
\label{LLMRLFRAMEWORKFIG}
\end{figure*}

\section{LLM-enhanced RL for LAENet}

%\subsection{Concept of LLM-enhanced RL}

%Introduce the definition, structure and characteristics of LLM-enhanced RL. 

Based on the above analysis, the concept of LLM-enhanced RL can be defined as the methods that integrate the high-level cognitive capabilities of LLMs, such as multimodal information processing, understanding, reasoning, planning, and generating, into the RL paradigm. Thus, we propose an LLM-enhanced RL framework for the LAENet, as shown in Fig. \ref{LLMRLFRAMEWORKFIG}, which leverages the complementary strengths of LLMs and RL to address the limitations of classical RL.

\subsection{Overview of LLM-enhanced RL Framework}

The LLM-enhanced RL framework for the LAENet leverages the high-level cognitive capabilities of LLMs to augment multiple stages of RL. It processes environmental states, generates actions, simulates outcomes, and shapes rewards with the support of LLMs while the RL agent iteratively learns optimal policies through continuous interaction, as shown in Fig. \ref{LLMRLFRAMEWORKFIG}. Thus, the classical RL loop with the support of LLMs ensures that the LAENet can handle dynamic, uncertain, and multimodal real-world scenarios with greater flexibility, collaboration, and generalization. Specifically, the four key roles of LLMs for enhancing RL in the LAENet are detailed as follows.

%Explain the different roles of LLM corresponding to issues in RL:

\subsubsection{LLM as Information Processor} As information processors, LLMs play a key role in decoupling the burden of interpreting complex, multimodal data from the RL agent. On the one hand, LLMs can extract meaningful features from raw observations using powerful pre-trained models. The aerial vehicles with generalization capabilities can quickly understand the changing environment (e.g., variations in communication conditions, weather, or terrain) and generate effective state representations for RL without retraining, as shown in Fig. \ref{LLMRLFRAMEWORKFIG}. For example, if an aerial vehicle's camera captures the weather changing from sunny to foggy, LLMs processes this input and outputs a compressed feature vector (e.g., ``visibility=low") as part of the RL state space. On the other hand, the LAENet with LLMs can reduce learning complexity for the RL agent by transferring informal natural language information (such as from ground control or user requests) into a formal task-specific language  \cite{10551749}. When ground control updates instructions (e.g., ``Emergency: Redirect to Zone $B$"), LLMs can interpret the urgency of redirect and update the RL agent’s objective to prioritize reaching Zone $B$.
%LLM help RL by processing complex multimodal inputs, translating them into structured task representations to ease policy learning and improve performance.

%The paper references work (e.g., Kwon et al., 2023) where LLMs convert safety guidelines (e.g., "Maintain altitude above 200m in urban areas") into constraint-aware reward terms:

\subsubsection{LLM as Reward Designer} As reward designers, LLMs can leverage extensive pre-trained knowledge and reasoning capabilities to shape and refine reward functions for RL agents. Specifically, LLMs can provide reward signals by generating task-relevant functions by interpreting descriptions and observations. Furthermore, LLMs serve as reward designers by generating executable reward function code, which delineates the calculation process and can be iteratively refined with feedback. Taking the task scheduling scenario in LAENet as an example, LLMs can generate task-specific reward functions by processing textual descriptions of objectives, such as ``minimize critical task delay", ``prioritize low-energy UAVs", and ``balance energy consumption and latency". Then, LLMs convert objectives into executable code snippets that compute reward signals during RL training. The feedback from the training process and changing environment can be used by LLMs to continuously refine the reward function, thereby overcoming the fixity and complexity of manually crafting rewards.
%LLM can assist RL in sparse-reward environments by providing implicit or explicit reward models through knowledge-driven reasoning and code generation.

\subsubsection{LLM as Decision-maker} LLMs act as decision-makers in RL by guiding or generating action sequences through their pre-trained knowledge, structured reasoning, and language understanding. As action-guiders, LLMs assist in reducing the action search space by generating high-quality action candidates based on task comprehension and contextual prompts. In action generation, LLMs are fine-tuned or prompted with task goals and trajectory histories to predict high-reward expert action policies. These candidates or expert actions can be incorporated and distilled into RL agents to regularize the policy learning. In scenarios where aerial vehicles need to optimize their trajectories to provide communication coverage for users, LLMs act as action-guiders by processing the current UAV state (e.g., location, nearby users, and energy levels) and generating a set of action candidates (e.g., “hover above user cluster A,” “move south 10 meters,” “move down 3 meters”). RL agents can select the most rewarding action using a value function, which may effectively avoid the large scale and noise in the action space of regular RL algorithms. Meanwhile, if prior UAV trajectory sequences resulted in incomplete communication coverage, LLMs can reason based on trajectory histories: “UAV is near a user cluster A; the unserved users are mostly north. It has 30\% battery. Therefore, it should head north at low speed” to generate actions.

%LLM enhance RL efficiency by generating or guiding actions using their semantic understanding and prior knowledge, improving both sample and exploration efficiency.

\subsubsection{LLM as Generator} LLMs can predict future state and reward sequences based on current observations and actions, which enables model-based RL agents to learn from simulated experiences. In UAV trajectory optimization tasks, states (e.g., UAV location, user distribution, and remaining energy) and actions (e.g., flight direction, and speed) can be input into the LLM to generate state-action-reward sequences. Subsequently, the LLM can produce large amounts of simulated trajectory data to support policy learning, thereby alleviating the limitation of classical RL’s reliance on interactions with environment. In addition, by prompting LLMs with decision paths of RL, LLMs can generate natural language to interpret policies to demonstrate their trust and transparency. The operators serving LAENet can trust the RL policy decision paths (e.g., UAV moved toward access point $k$ to serve user $n$) since the LLM can generate interpretable descriptions “UAV chose to move east to minimize energy cost and meet the imminent offloading deadline of user $n$.”

%LLMs support model-based and explainable RL by generating accurate world models for planning and providing interpretable policy explanations through reasoning and multimodal understanding.

%\subsection{Application of LLM-enhanced RL for LAENet}

\begin{figure*}[t]
\centering
\includegraphics[width=1\linewidth]{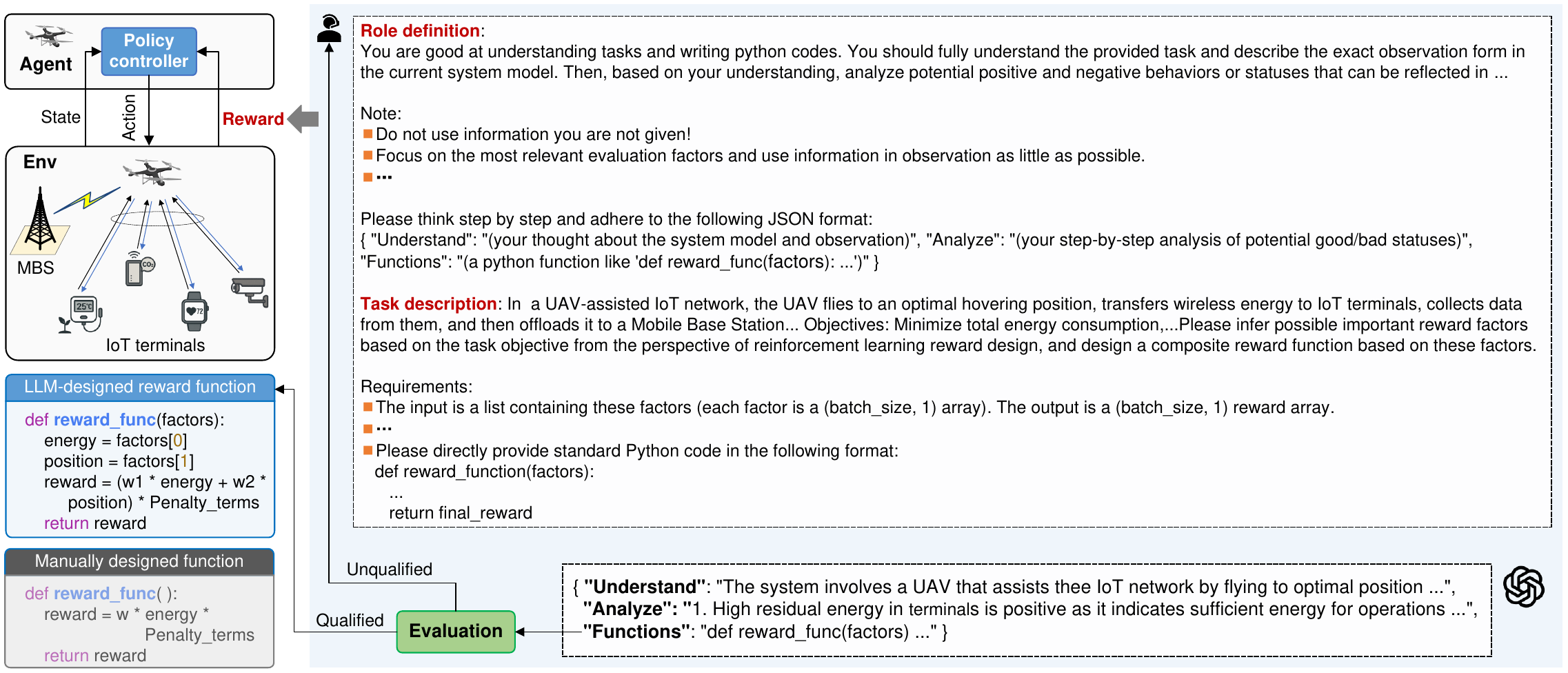}
\caption{UAV-assisted IoT network with LLM-designed reward funtioon for RL in the LAENet. The UAV agent interacts with the environment by selecting actions based on observed states. The LLM generates the reward function based on structured prompt input of role definition and task description. The generated reward function is evaluated through predefined constraints before being applied to policy learning.}
\label{LLMASREWARDSYSTEMMODEL}
%\vspace{-0.5cm}
\end{figure*}

\subsection{Workflow of LLM-enhanced RL framework for LAENet}

% 注意！！！这里替换成不同的应用场景更好，结合我综述写的。下面的这些细分可以用于future research directions
%The functions in LAENet, such as task scheduling \cite{9397778}, resource allocation \cite{10102429}, computation offloading \cite{10445426}, distributed learning \cite{10334003}, trajectory optimization \cite{10239498}, can be improved by using LLM-enhanced RL from the aspects of information processor, reward designer, decision-maker, and generator.

We take UAV-assisted data collection for the Internet of Things (IoT) in LAENet as an example to illustrate the workflow of LLM-enhanced RL framework in Fig. \ref{LLMRLFRAMEWORKFIG}.

{\bf Step 1: State Perception and Abstraction.} The interaction between the UAV and the environment is modeled as a Markov Decision Process (MDP)\footnote{MDP is a mathematical framework for modeling decision-making situations where outcomes are partly random and partly under the control of a decision-maker. It consists of states, actions, transition probabilities, and rewards.\label{s1}}, where each state captures spatial, energy, and communication conditions relevant to decision-making. Leveraging the capabilities of LLMs, the UAV abstracts natural language instructions or sensor descriptions (e.g., a command such as ``capture aerial images of a congested area" or a sensor report indicating ``battery level is low at terminal $n$") into compact and informative state representations.

{\bf Step 2: Action Selection and Policy Execution.} Based on the current state and the learned policy, the agent generates actions to optimize long-term objectives (e.g., minimizing energy consumption) by prompting and guiding the LLM to reason about adjustments to flight paths or scheduling of data collection. This process can be governed by various methods of reinforcement learning, including policy-based approaches\footnote{Policy-based methods directly learn a mapping from states to actions (i.e., a policy), such as Deep Deterministic Policy Gradient (DDPG) and Twin Delayed Deep Deterministic Policy Gradient (TD3).\label{s2}}, value-based methods\footnote{Value-based methods estimate the expected return (value) of taking an action in a given state, and derive the policy from these values. Common examples include Q-learning and Deep Q-Networks (DQN).\label{s3}}, and model-based techniques\footnote{Model-based techniques learn a model of the environment's dynamics (i.e., transition probabilities) and use this model for planning or policy learning, such as in Dyna-Q or Model Predictive Control (MPC).\label{s4}}.

%Proximal Policy Optimization (PPO), 

{\bf Step 3:  Reward Evaluation and Feedback Processing.} After executing an action, the agent receives an informative and adaptive reward shaped by the LLM to quantify performance with respect to predefined objectives. For example, the user may say ``I am happy with the service speed" which means the cost in terms of delay is low. These enriched reward signals guide the agent toward more effective optimization and better respond to system constraints.

{\bf Step 4:  Policy Update and Knowledge Integration.} Based on accumulated experience, the agent uses the LLM to update and refine its policy by integrating external knowledge (such as statistical information on terminal data transmission patterns or communication conditions in certain areas of UAV hovering), enabling generalization across tasks, and assisting in the interpretation of policy behaviors to support human-in-the-loop optimization.

Embedding LLMs into each stage of the RL loop enables agents to operate more intelligently and adaptively in complex, dynamic environments. This integration enhances learning efficiency, improves generalization to unknown scenarios, and facilitates human-aligned decision-making in the LAENet.

%\subsection{Applications on LAENet}

%Review or propose the application of LLM-enhanced RL as the information processor, reward designer, decision-maker and generator for LAENet 

%\section{LLM-enhanced RL for LAENet}

%\subsection{Framework Overview}

%LLM-enhanced RL is used for collaborative computing framework in LAENet. For example, when mul-UAV face multiple tasks, they need to dynamically decide how to perform task scheduling, whether to offload to other UAVs or edge servers, and how to allocate resources. LLM-enhanced state representation or decision-maker is very beneficial.

%\subsection{Use Cases}

\section{Case study: LLM as reward designer to enhance RL for energy optimization in LAENet}

%1页-1.25页

%LLM-enhanced RL enables UAVs to more intelligently formulate optimal trajectory optimization strategies through interaction with the environment, thereby improving the performance and efficiency of LAENet.

\subsection{System Overview}

%We consider a UAV-assisted IoT scenario in the LAENet, comprising a UAV, a macro base station (MBS), and multiple distributed IoT terminals, as shown in Fig. \ref{LLMASREWARDSYSTEMMODEL}. The UAV operates at a fixed altitude and speed and performs two main functions: wirelessly charging IoT terminals via energy transfer and collecting data from terminals. The UAV dynamically selects its hovering position based on terminal locations or channel conditions. The IoT terminals harvest energy to transmit data to the UAV, which subsequently relays aggregated data to the MBS. The Age of Information (AoI) metric is adopted to characterize data freshness by tracking delays between data generation and successful decoding. The total energy consumption includes the transmission energy of all IoT terminals and the UAV’s energy consumption. The energy consumption of the UAV comprises propulsion (flight and hovering) and communication (data collection and offloading). The optimization problem is formulated to minimize total system energy subject to constraints on UAV positioning, power limits, data throughput, decoding reliability, and AoI thresholds.

We consider a UAV-assisted IoT scenario in LAENet, involving a UAV, a macro base station (MBS), and multiple distributed IoT terminals, as shown in Fig. \ref{LLMASREWARDSYSTEMMODEL}. The UAV flies at fixed altitude and constant speed to dynamically hover near IoT terminals. The terminals utilize energy harvested from the UAV to transmit data, which is subsequently aggregated by the UAV and relayed to the MBS. The system’s total energy consumption includes the transmission energy of terminals, the propulsion and communication energy of the UAV. The objective is to minimize total energy under constraints on power limits, data throughput, decoding reliability, and data freshness.

\subsection{Implementation Details of LLM as Reward Designer}

%In this part, we provide a detailed introduction to the LLM as reward designer module, which leverages the prior knowledge and strong reasoning capabilities of LLMs to generate the executable reward function.

We provide a detailed implementation of the LLM as reward designer module for the LAENet

%LLMs need to be guided to think in the role of reward designers in RL, ensuring they can understand the task in the environment, the rules for designing rewards, and the coding ability for reward function generation. Based on chain-of-thought techniques \cite{NEURIPS2022_9d560961} and appropriate prompt design \cite{karmaker-santu-feng-2023-teler}, we guide the LLMs through two aspects: {\color{blue}the prompt design of role definition and task expression. As shown in Fig. \ref{LLMASREWARDSYSTEMMODEL}, The role definition specifies the workflow in which the LLM serves as a reward designer, including instructing the LLM to use Python programming, comprehend the system model of the task, reason based on observations, and generate the reward function accordingly. Additionally, a set of Notes is provided to the LLM as fundamental normative constraints. The output format of the LLM’s response to the prompts (e.g., a JSON format that is easy to parse) is also standardized.} Task expression involves describing the task objectives and the behaviors of RL agents to the LLM, while also guiding the LLM to extract the structure, features, and meanings of states and actions. These prompts help the LLM understand task-relevant contextual information, which avoids overly generalized responses and reduces the burden of prompt engineering across tasks.

\subsubsection{User Prompt Design}The users need to guide LLMs to think in the role of reward designers in RL, ensuring that LLMs can understand the task in the environment, the rules for designing rewards, and the coding ability for reward function generation. Therefore, the users are required to provide effective prompts to the LLMs. Based on chain-of-thought techniques \cite{NEURIPS2022_9d560961} and appropriate prompt design \cite{karmaker-santu-feng-2023-teler}, we propose a guideline for users' prompts which are divided into two elements: role definition and task description. 

{\bf Role Definition:} As shown in Fig. \ref{LLMASREWARDSYSTEMMODEL}, the role definition has three parts. The first part specifies the functions of the LLM as a reward designer, including comprehending the system model of the task, reasoning based on observations, and using Python programming to generate the reward function accordingly. The second part includes a set of notes for the LLM as fundamental normative constraints, such as not using ungiven information and focusing on the most relevant factors. The third part standardizes the output format of the LLM’s response to the prompts, i.e., a JSON format that is easy to parse. 

{\bf Task description: }Task description defines the system model and optimization objective of a specific scenario, which helps the LLM understand task-relevant contextual information, avoids overly generalized responses, and reduces the burden of prompt engineering across tasks. It also requires the LLM to infer important reward factors in designing the reward function, which can effectively guide the RL agent to optimize the actions in the learning process. Additionally, the input and output formats for factors and the reward function are standardized.

%原大小：fig4 0.9 fig5 0.8

\begin{figure}[t]
\centering
\includegraphics[width=0.95\linewidth]{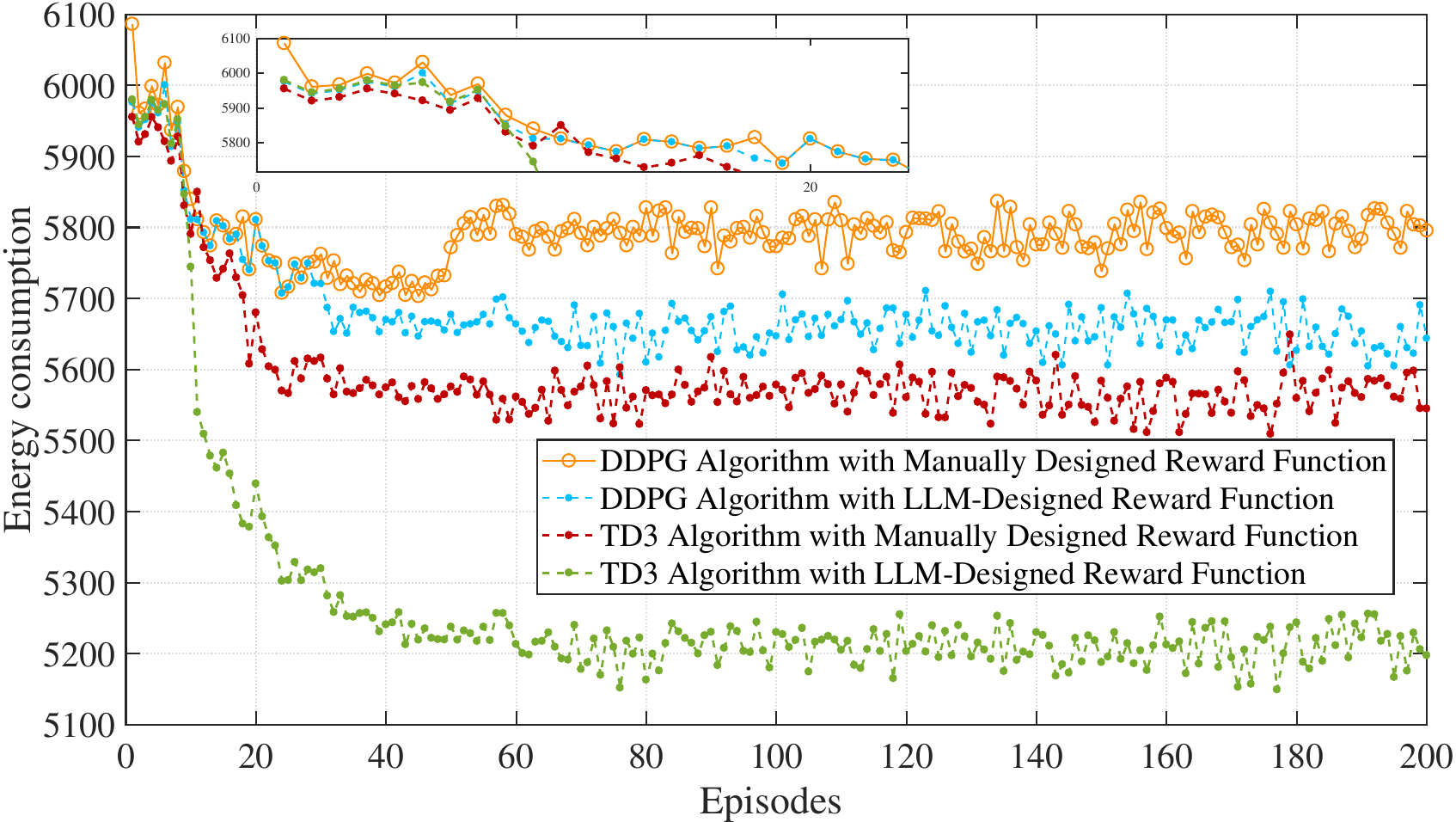}
\caption{Energy consumption over episodes of different algorithm with manually designed and LLM-generated reward functions.}
\label{DDPGTD3MANUANDLLM}
\end{figure}

\begin{figure}t]
\centering
\includegraphics[width=0.9\linewidth]{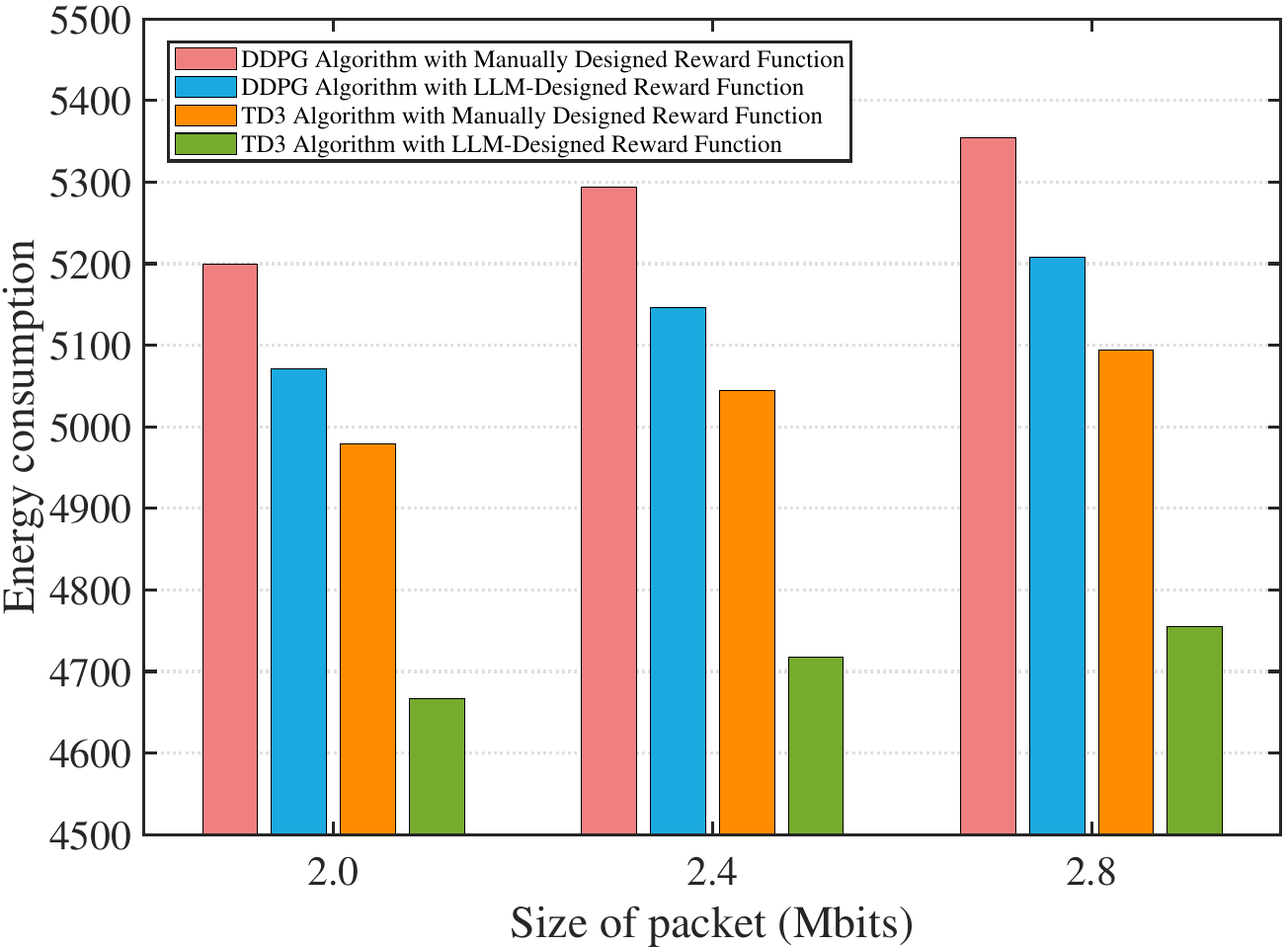}
\caption{Effect of packet size on energy consumption using different reward design methods. }
\label{energywithdiffpacketmaullm}
\end{figure}

\subsubsection{LLM Response for Reward Design} The LLM receives the prompts from the user and generates the reward function through code generation ability along with logical reasoning. However, the response of the LLM is stochastic due to its probabilistic nature \cite{yang2024rl2}. In addition, the LLM may hallucinate and generate code that appears reasonable but is actually non-executable \cite{dalal2024planseqlearn}. 

{\bf LLM-designed Reward Evaluation:} Inspired by the recent work \cite{NEURIPS2023_1b44b878}, the LLM is required to generate multiple candidate reward functions rather than relying on a single random response. Specifically, each candidate function is generated by prompting the LLM using the role definition and task expression to reflect from a logical consistency perspective, until the reward function is evaluated to satisfy the constraints. The constraints include whether the LLM response is successful, whether the output is in a valid JSON structure, and whether the return type of the reward function is correct. 

{\bf Exploration of Reward Factors:} As shown in Fig. \ref{LLMASREWARDSYSTEMMODEL}, our proposed LLM-assisted reward function (i.e., $reward = (w_1 \times energy + w_2 \times position) \times Penalty$), compared to the manually designed function (i.e., $reward = w \times energy \times Penalty$), further considers the UAV's selection of more optimal positions to contribute to minimizing the total energy. This reward factor helps reduce propulsion energy by encouraging the UAV to stay closer to the center of the sensor distribution to reduce flight distance and travel time. Additionally, being near the center can indirectly save energy due to faster data transmission and shorter collection periods, which reduce the time the UAV needs to hover and communicate.
%Then, the role definition, task expression, and candidate reward functions are used to re-prompt the LLM to reflect from a logical consistency perspective to obtain the final reward function. Finally, before applying the generated reward function to RL, its executability is evaluated based on relevant pre-defined constraints, such as whether the penalty coefficient for aircraft collisions falls within a valid range, and whether the reward function takes the form of a linear combination.

%这些候选函数是通过利用role definition, task expression  to  prompt the LLM to reflect from a logical consistency perspective to generate reward function. 简要说明用什么约束来验证函数正确性

\subsection{Performance Evaluation}

We validate the superiority of using LLMs as reward designers for RL in the LAENet. Two DRL algorithms, DDPG and TD3, are adopted to conduct the simulations. The actor and critic networks are trained with learning rates of $10^{-4}$ and $3 \times 10^{-4}$, respectively. A batch size of 64, training episodes of 200, and a discount factor of 0.99 are set. The simulation environment consists of a 300m $ \times $ 300m square area representing a marine IoT coverage zone, where 10 IoT terminals are randomly deployed. The wireless channel between the UAV and the terminals follows the Rician fading model. We use the reward design method from previous work \cite{9750860} as a baseline, namely manually designed reward functions. As shown in Fig. \ref{LLMASREWARDSYSTEMMODEL}, the manually designed reward function includes energy-related reward terms and penalty terms. In contrast, we employ GPT-4o as the LLM module to design the reward function, which incorporates richer reward factors based on the position of the UAV.

Fig. \ref{DDPGTD3MANUANDLLM} shows the convergence performance of DDPG and TD3 algorithms using manually designed and LLM-generated reward functions. It can be observed that algorithms with LLM-designed rewards consistently outperform their manually designed functions in reducing energy consumption, with TD3 algorithm achieving up to 7.2\% lower final energy consumption. The reward structure designed by the LLM encourages the UAV to select more efficient trajectories and reduce flight and communication overhead. This improvement can be attributed to the LLM’s ability to incorporate high-level reasoning and task-specific context when generating reward functions. 
%These results validate the effectiveness of leveraging LLMs for reward function generation, especially when combined with advanced RL algorithms like TD3.

Fig. \ref{energywithdiffpacketmaullm} shows the impact of varying packet sizes on energy consumption. As packet size increases from 2.0 to 2.8 Mbits, overall energy consumption also rises, due to the prolonged data collection and transmission periods required for larger packets. It is evident that algorithms guided by LLM-designed reward functions consistently outperform those using manually crafted rewards, especially achieving up to 6.2\% lower energy consumption at the 2.0 Mbits packet size, which shows the effectiveness of our LLM-guided reward design in optimizing UAV decision-making and reducing system energy overhead.

%\section{Future Research Directions}

%0.5页

\section{Conclusion and Future Work}

In this paper, we have explored the integration of LLMs into RL to address key challenges in the LAENet. By leveraging the strengths of LLMs, we have proposed an LLM-enhanced RL framework to mitigate limitations throughout the entire pipeline of classical RL. Finally, we have presented a case study to demonstrate the effectiveness of using LLMs for designing reward functions. The promising directions from this work include the development of modular LLM-RL agents with specialized capabilities, such as planning, memory, tool use, and retrieval-augmented reasoning, to enable more adaptive and context-aware decision-making. Furthermore, in multi-agent RL scenarios, multiple collaborative LLMs can assume complementary roles, opening up new possibilities for addressing complex and dynamic tasks in heterogeneous, resource-constrained environments. Advancing these directions should be critical for realizing intelligent, efficient, and scalable aerial networking systems in the LAENet and other real-world applications.

\bibliographystyle{IEEEtran}
\bibliography{ref}

\begin{thebibliography}{10}
\providecommand{\url}[1]{#1}
\csname url@samestyle\endcsname
\providecommand{\newblock}{\relax}
\providecommand{\bibinfo}[2]{#2}
\providecommand{\BIBentrySTDinterwordspacing}{\spaceskip=0pt\relax}
\providecommand{\BIBentryALTinterwordstretchfactor}{4}
\providecommand{\BIBentryALTinterwordspacing}{\spaceskip=\fontdimen2\font plus
\BIBentryALTinterwordstretchfactor\fontdimen3\font minus \fontdimen4\font\relax}
\providecommand{\BIBforeignlanguage}[2]{{%
\expandafter\ifx\csname l@#1\endcsname\relax
\typeout{** WARNING: IEEEtran.bst: No hyphenation pattern has been}%
\typeout{** loaded for the language `#1'. Using the pattern for}%
\typeout{** the default language instead.}%
\else
\language=\csname l@#1\endcsname
\fi
#2}}
\providecommand{\BIBdecl}{\relax}
\BIBdecl

\bibitem{10759668}
Z.~Li \emph{et~al.}, ``Unauthorized {UAV} countermeasure for low-altitude economy: Joint communications and jamming based on {MIMO} cellular systems,'' \emph{IEEE Internet Things J.}, vol.~12, no.~6, pp. 6659--6672, 2025.

\bibitem{10972017}
Q.~Wei \emph{et~al.}, ``Multi-{UAV}-enabled energy-efficient data delivery for low-altitude economy: Joint coded caching, user grouping, and {UAV} deployment,'' \emph{IEEE Internet Things J.}, pp. 1--1, 2025.

\bibitem{9750860}
O.~S. Oubbati \emph{et~al.}, ``Synchronizing {UAV} teams for timely data collection and energy transfer by deep reinforcement learning,'' \emph{IEEE Trans. Veh. Technol.}, vol.~71, no.~6, pp. 6682--6697, 2022.

\bibitem{10766898}
Y.~Cao \emph{et~al.}, ``Survey on large language model-enhanced reinforcement learning: Concept, taxonomy, and methods,'' \emph{IEEE Trans. Neural Netw. Learn. Syst.}, pp. 1--21, 2024.

\bibitem{10679152}
R.~Zhang \emph{et~al.}, ``Generative {AI} agents with large language model for satellite networks via a mixture of experts transmission,'' \emph{IEEE J. Sel. Areas Commun.}, vol.~42, no.~12, pp. 3581--3596, 2024.

\bibitem{NEURIPS2022_9d560961}
J.~Wei \emph{et~al.}, ``Chain-of-thought prompting elicits reasoning in large language models,'' in \emph{Proc. NeurIPS}, vol.~35, 2022, pp. 24\,824--24\,837.

\bibitem{kwon2023reward}
M.~Kwon \emph{et~al.}, ``Reward design with language models,'' in \emph{Proc. ICLR}, 2023.

\bibitem{pmlr-v162-paischer22a}
F.~Paischer \emph{et~al.}, ``History compression via language models in reinforcement learning,'' in \emph{Proc. ICML}, 2022, pp. 17\,156--17\,185.

\bibitem{dalal2024planseqlearn}
M.~Dalal \emph{et~al.}, ``Plan-seq-learn: Language model guided {RL} for solving long horizon robotics tasks,'' in \emph{Proc. ICLR}, 2024.

\bibitem{shi2024unleashing}
R.~Shi \emph{et~al.}, ``Unleashing the power of pre-trained language models for offline reinforcement learning,'' in \emph{Proc. ICLR}, 2024.

\bibitem{NEURIPS2023_1b44b878}
N.~Shinn \emph{et~al.}, ``Reflexion: language agents with verbal reinforcement learning,'' in \emph{Proc. NeurIPS}, vol.~36, 2023, pp. 8634--8652.

\bibitem{yang2024rl2}
X.~Yang \emph{et~al.}, ``Rl2: Reinforce large language model to assist safe reinforcement learning for energy management of active distribution networks,'' \emph{arXiv preprint arXiv:2412.01303}, 2024.

\bibitem{10529514}
L.~Chen \emph{et~al.}, ``Rlingua: Improving reinforcement learning sample efficiency in robotic manipulations with large language models,'' \emph{IEEE Robot. Autom. Lett.}, vol.~9, no.~7, pp. 6075--6082, 2024.

\bibitem{10551749}
Y.~Han \emph{et~al.}, ``Large language model guided reinforcement learning based six-degree-of-freedom flight control,'' \emph{IEEE Access}, vol.~12, pp. 89\,479--89\,492, 2024.

\bibitem{karmaker-santu-feng-2023-teler}
K.~Santu \emph{et~al.}, ``{TEL}e{R}: A general taxonomy of {LLM} prompts for benchmarking complex tasks,'' in \emph{Findings Assoc. Comput. Linguist.: EMNLP}, 2023, pp. 14\,197--14\,203.

\end{thebibliography}

\vfill

\end{document}